\documentclass[11pt,a4paper]{article}
\usepackage[nohyperref]{acl2017}
\usepackage{times}
\usepackage{latexsym}
\usepackage{xspace}
\usepackage{url}
\usepackage{graphicx}
\usepackage{amsmath}
\usepackage{amssymb}
\usepackage{tabularx}
\usepackage{float}

\aclfinalcopy % Uncomment this line for the final submission
\newcommand{\mdcnn}{McDepCNN\xspace}

\title{Deep learning for extracting protein-protein interactions from biomedical literature}

\author{Yifan Peng \hspace*{5em} Zhiyong Lu\\
  National Center for Biotechnology Information \\
  National Library of Medicine \\
  National Institutes of Health \\
  Bethesda, MD 20894 \\
  {\tt \{yifan.peng, zhiyong.lu\}@nih.gov}}

\date{}

\begin{document}
\maketitle
\begin{abstract}
State-of-the-art methods for protein-protein interaction (PPI) extraction are primarily feature-based or kernel-based by leveraging lexical and syntactic information. But how to incorporate such knowledge in the recent deep learning methods remains an open question. In this paper, we propose a multichannel dependency-based convolutional neural network model (\mdcnn). It applies one channel to the embedding vector of each word in the sentence, and another channel to the embedding vector of the head of the corresponding word. Therefore, the model can use richer information obtained from different channels. Experiments on two public benchmarking datasets, AIMed and BioInfer, demonstrate that \mdcnn compares favorably to the state-of-the-art rich-feature
and single-kernel based methods. In addition, \mdcnn achieves 24.4\% relative improvement in F1-score over the state-of-the-art methods on cross-corpus evaluation and 12\% improvement in F1-score over kernel-based methods on ``difficult'' instances. These results suggest that \mdcnn generalizes more easily over different corpora, and is capable of capturing long distance features in the sentences.
\end{abstract}

\section{Introduction}

With the growing amount of biomedical information available in the textual form, there has been considerable interest in applying natural language processing (NLP) techniques and machine learning (ML) methods to the biomedical literature~\cite{huang2015community,leaman2016taggerone,singhal2016text,peng2016improving}. One of the most important tasks is to extract protein-protein interaction relations~\cite{krallinger2008overview}.

Protein-protein interaction (PPI) extraction is a task to identify interaction relations between protein entities mentioned within a document. While PPI relations can span over sentences and even cross documents, current works mostly focus on PPI in individual sentences~\cite{pyysalo2008comparative,tikk2010comprehensive}. For example, ``ARFTS'' and ``XIAP-BIR3'' are in a PPI relation in the sentence ``ARFTS$_{\text{PROT1}}$ specifically binds to a distinct domain in XIAP-BIR3$_{\text{PROT2}}$''. 

Recently, deep learning methods have achieved notable results in various NLP tasks~\cite{manning2015computational}. For PPI extraction, convolutional neural networks (CNN) have been adopted and applied effectively~\cite{zeng2014relation, quan2016multichannel, hua2016shortest}. Compared with traditional supervised ML methods, the CNN model is more generalizable and does not require tedious feature engineering efforts. However, how to incorporate linguistic and semantic information into the CNN model remains an open question. Thus previous CNN-based methods have not achieved state-of-the-art performance in the PPI task~\cite{zhao2016protein}.

In this paper, we propose a multichannel dependency-based convolutional neural network, \mdcnn, to provide a new way to model the syntactic sentence structure in CNN models. Compared with the widely-used one-hot CNN model (e.g., the shortest-path information is firstly transformed into a binary vector which is zero in all positions except at this shortest-path's index, and then applied to CNN), \mdcnn utilizes a separate channel to capture the dependencies of the sentence syntactic structure.

To assess \mdcnn, we evaluated our model on two benchmarking PPI corpora, AIMed~\cite{bunescu2005comparative} and BioInfer~\cite{pyysalo2007bioinfer}. Our results show that \mdcnn performs better than the state-of-the-art feature- and kernel-based methods. 

We further examined \mdcnn in two experimental settings: a cross-corpus evaluation and an evaluation on a subset of ``difficult'' PPI instances previously reported~\cite{tikk2013detailed}. Our results suggest that \mdcnn is more generalizable and capable of capturing long distance information than kernel methods.

The rest of the manuscript is organized as follows. We first present related work. Then, we describe our model in Section~\ref{sec:cnn}, followed by an extensive evaluation and discussion in Section~\ref{sec:results}. We conclude in the last section.

\section{Related work}

From the ML perspective, we formulate the PPI task as a binary classification problem where discriminative classifiers are trained with a set of positive and negative relation instances. In the last decade, ML-based methods for the PPI tasks have been dominated by two main types: the feature-based vs. kernel based method. The common characteristic of these methods is to transform relation instances into a set of features or rich structural representations like trees or graphs, by leveraging linguistic analysis and knowledge resources. Then a discriminative classifier is used, such as support vector machines~\cite{vapnik1995nature} or conditional random fields~\cite{lafferty2001conditional}.

While these methods allow the relation extraction systems to inherit the knowledge discovered by the NLP community for the pre-processing tasks, they are highly dependent on feature engineering~\cite{fundel2007relex, vanlandeghem2008extracting, miwa2009rich, bui2011ahybrid}. The difficulty with feature-based methods is that data cannot always be easily represented by explicit feature vectors. 

Since natural language processing applications involve structured representations of the input data, deriving good features is difficult, time-consuming, and requires expert knowledge. Kernel-based methods attempt to solve this problem by implicitly calculating dot products for every pair of examples~\cite{erkan2007semi, airola2008all, miwa2009protein, kim2010walk, chowdhury2011astudy}. Instead of extracting feature vectors from examples, they apply a similarity function between examples and use a discriminative method to label new examples~\cite{tikk2010comprehensive}. However, this method also requires manual effort to design a similarity function which can not only encode linguistic and semantic information in the complex structures but also successfully discriminate between examples. Kernel-based methods are also criticized for having higher computational complexity~\cite{collins2002new}.

Convolutional neural networks (CNN) have recently achieved promising results in the PPI task~\cite{zeng2014relation, hua2016shortest}. CNNs are a type of feed-forward artificial neural network whose layers are formed by a convolution operation followed by a pooling operation~\cite{lecun1998gradient}. Unlike feature- and kernel-based methods which have been well studied for decades, few studies investigated how to incorporate syntactic and semantic information into the CNN model. To this end, we propose a neural network model that makes use of automatically learned features (by different CNN layers) together with manually crafted  ones (via domain knowledge), such as words, part-of-speech tags, chunks, named entities, and dependency graph of sentences. Such a combination in feature engineering has been shown to be effective in other NLP tasks also (e.g.~\cite{shimaoka2017neural}). 

Furthermore, we propose a multichannel CNN, a model that was suggested to capture different ``views'' of input data. In the image processing, \cite{krizhevsky2012imagenet} applied different RGB (red, green, blue) channels to color images. In NLP research, such models often use separate channels for different word embeddings~\cite{yin2015multichannel, shi2016multichannel}. For example, one could have separate channels for different word embeddings~\cite{quan2016multichannel}, or have one channel that is kept static throughout training and the other that is fine-tuned via backpropagation~\cite{kim2014convolutional}. Unlike these studies, we utilize the head of the word in a sentence as a separate channel.

\section{CNN Model}
\label{sec:cnn}

\subsection{Model Architecture Overview}
\begin{figure*}
\centering
%\fbox{%
\includegraphics[width=.98\textwidth,trim={.5cm 4cm 22cm 0}]{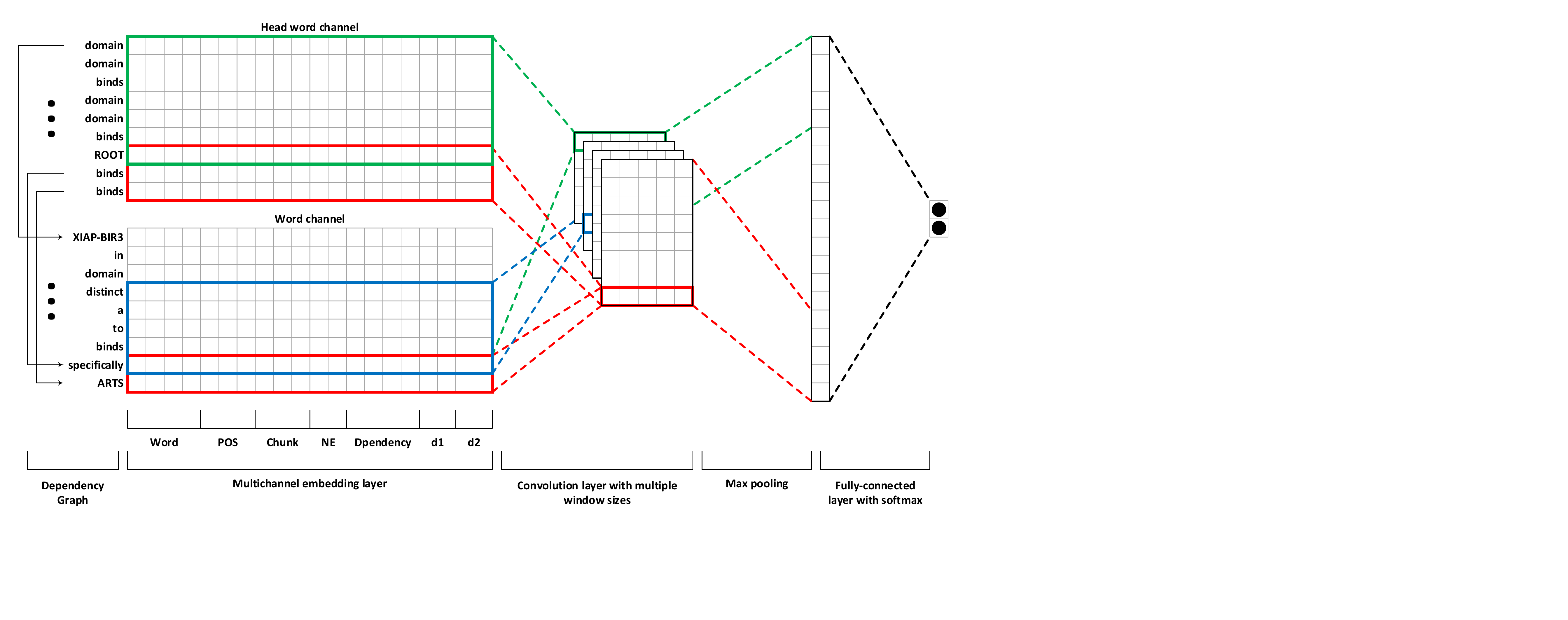}
%}
\caption{Overview of the CNN model.\label{fig:overview}}
\end{figure*}

Figure~\ref{fig:overview} illustrates the overview of our model, which takes a complete sentence with mentioned entities as input and outputs a probability vector (two elements) corresponding to whether there is a relation between the two entities. Our model mainly consists of three layers: a multichannel embedding layer, a convolution layer, and a fully-connected layer.

\subsection{Embedding Layer}

In our model, as shown in Figure~\ref{fig:overview}, each word in a sentence is represented by concatenating its word embedding, part-of-speech, chunk, named entity, dependency, and position features.

\subsubsection{Word embedding}
Word embedding is a language modeling techniques where words from the vocabulary are mapped to vectors of real numbers. It has been shown to boost the performance in NLP tasks. In this paper, we used pre-trained word embedding vectors~\cite{pyysalo2013distributional} learned on PubMed articles using the word2vec tool~\cite{mikolov2013distributed}. The dimensionality of word vectors is 200.

\subsubsection{Part-of-speech}
We used the part-of-speech (POS) feature to extend the word embedding. Similar to~\cite{zhao2016drug}, we divided POS into eight groups. Then each group is mapped to an eight-bit binary vector. In this way, the dimensionality of the POS feature is 8.

\subsubsection{Chunk}
We used the chunk tags obtained from Genia Tagger for each word~\cite{tsuruoka2005bidirectional}. We encoded the chunk features using a one-hot scheme. The dimensionality of chunk tags is 18.

\subsubsection{Named entity}
To generalize the model, we used four types of named entity encodings for each word. The named entities were provided as input by the task data. In one PPI instance, the types of two proteins of interest are PROT1 and PROT2 respectively. The type of other proteins is PROT, and the type of other words is O. If a protein mention spans multiple words, we marked each word with the same type (we did not use a scheme such as IOB). The dimensionality of named entity is thus 4.

\subsubsection{Dependency}
\begin{figure}
\centering
\includegraphics[width=.49\textwidth]{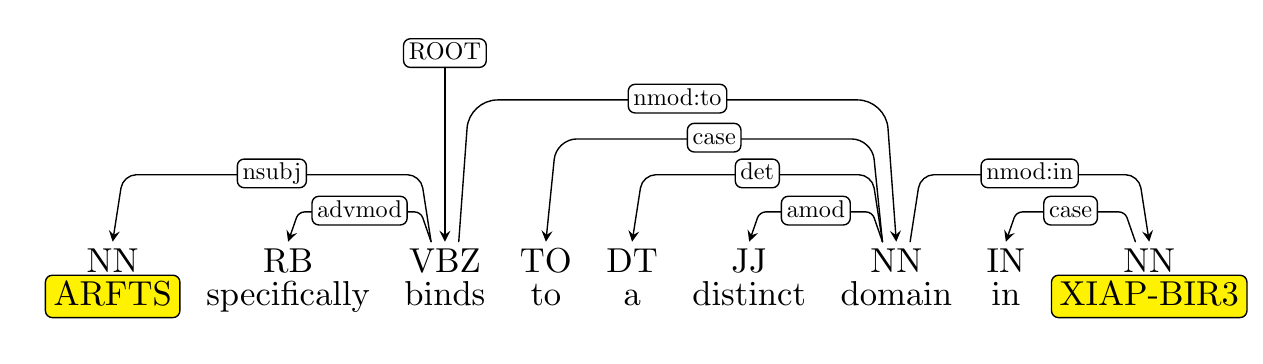}
\caption{Dependency graph.\label{fig:dg}}
\end{figure}

To add the dependency information of each word, we used the label of ``incoming'' edge of that word in the dependency graph. Take the sentence from Figure~\ref{fig:dg} as an example, the dependency of ``ARFTS'' is ``nsubj'' and the dependency of ``binds'' is ``ROOT''. We encoded the dependency features using a one-hot scheme, and their dimensionality is 101.

\subsubsection{Position feature}
In this work, we consider the relationship of two protein mentions in a sentence. Thus, we used the position feature proposed in~\cite{sahu2016relation}, which consists of two relative distances, $d1$ and $d2$, for representing the distances of the current word to PROT1 and PROT2 respectively. For example in Figure~\ref{fig:dg}, the relative distances of the word ``binds'' to PROT1 (``ARFTs'') and PROT2 (``XIAP-BIR3'') are 2 and -6, respectively. Same as in Table S4 of~\cite{zhao2016drug}, both $d1$ and $d2$ are non-linearly mapped to a ten-bit binary vector, where the first bit stands for the sign and the remaining bits for the distance. 

\subsection{Multichannel Embedding Layer}

A novel aspect of \mdcnn is to add the ``head'' word representation of each word as the second channel of the embedding layer. For example, the second channel of the sentence in Figure~\ref{fig:dg} is ``binds binds ROOT binds domain domain binds domain'' as shown in Figure~\ref{fig:overview}. There are several advantages of using the ``head'' of a word as a separate channel. 

First, it intuitively incorporates the dependency graph structure into the CNN model.  Compared with~\cite{hua2016shortest} which used the shortest path between two entities as the sole input for CNN, our model does not discard information outside the scope of two entities. Such information was reported to be useful~\cite{zhou2007tree}. Compared with~\cite{zhao2016drug} which used the shortest path as a bag-of-word sparse 0-1 vector, our model intuitively reflects the syntactic structure of the dependencies of the input sentence. 

Second, together with convolution, our model can better capture longer distance dependencies than the sliding window size. As shown in Figure~\ref{fig:dg}, the second channel of \mdcnn breaks the dependency graph structure into structural $<$head word, child word$>$ pairs where each word is a modifier of its previous word. In this way, it reflects the skeleton of a constituent where the second channel shadows the detailed information of all sub-constituents in the first channel. From the perspective of the sentence string, the second channel is similar to a gapped $n$-gram or a skipped $n$-gram where the skipped words are based on the structure of the sentence. 

\subsection{Convolution}

We applied convolution to input sentences to combine two channels and get local features~\cite{collobert2011natural}. Consider $x_1,\dotsc,x_n$ to be the sequence of word representations in a sentence where 
\begin{equation}
x_i=E_{word}\oplus\dotsb\oplus E_{poistion}, i=1,\dotsc,n
\end{equation}
Here $\oplus$ is concatenation operation so $x^i \in \mathbb{R}^d$ is the embedding vector for the $i$th word with the dimensionality $d$. Let $x_{i:i+k-1}^{c}$ represent a window of size $k$ in the sentence for channel $c$. Then the output sequence of the convolution layer is
\begin{equation}
con_i=f(\sum_{c}{w_k^{c} x_{i:i+k-1}^{c}}+b_k)
\end{equation}
where $f$ is a rectify linear unit (ReLU) function and $b_k$ is the biased term. Both $w_k^{c}$ and $b_k$ are the learning parameters. 

1-max pooling was then performed over each map, i.e., the largest number from each feature map was recorded. In this way, we obtained fixed length global features for the whole sentence. The underlying intuition is to consider only the most useful feature from the entire sentence. 
\begin{equation}
m_k = \max_{1\leq i \leq n-k+1}{(con_i)}
\end{equation}

\subsection{Fully Connected Layer with Softmax}

To make a classifier over extracted global features, we first applied a fully connected layer to the feature vectors of multichannel obtained above. 
\begin{equation}
O = w_o (m_3 \oplus m_5 \oplus m_7) + b_o
\end{equation}

The final softmax then receives this vector $O$ as input and uses it to classify the PPI; here we assume binary classification for the PPI task and hence depict two possible output states.
\begin{equation}
%p(ppi|x,\theta)=\frac{\exp{(O_{ppi})}}{\exp{O_{ppi}}+\exp{O_{other}}}
p(ppi|x,\theta)=\frac{e^{O_{ppi}}}{e^{O_{ppi}}+e^{O_{other}}}
\end{equation}
Here, $\theta$ is a vector of the hyper-parameters of the model, such as $w_k^c$, $b_k$, $w_o$, and $b_o$. Further, we used dropout technique in the output of the max pooling layer for regularization~\cite{srivastava2014dropout}. This prevented our method from overfitting by randomly ``dropping'' with probability $(1-p)$ neurons during each forward/backward pass while training.

\subsection{Training}
To train the parameters, we used the log-likelihood of parameters on a mini-batch training with a batch size of $m$. We use the Adam algorithm to optimize the loss function~\cite{kingma2015adam}.
\begin{equation}
J(\theta) = \sum_{m}{p(ppi^{(m)}|x^{(m)},\theta})
\end{equation}

\subsection{Experimental setup}

For our experiments, we used the Genia Tagger to obtain the part-of-speech, chunk tags, and named entities of each word~\cite{tsuruoka2005bidirectional}. We parsed each sentence using the Bllip parser with the biomedical model~\cite{charniak2000maximum, mcclosky2009any}. The universal dependencies were then obtained by applying the Stanford dependencies converter on the parse tree with the \textit{CCProcessed} and \textit{Universal} options~\cite{de2014universal}.

We implemented the model using TensorFlow~\cite{abadi2016tensorflow}. All trainable variables were initialized using the Xavier algorithm~\cite{glorot2010understanding}. We set the maximum sentence length to 160. That is, longer sentences were pruned, and shorter sentences were padded with zeros. We set the learning rate to be 0.0007 and the dropping probability 0.5. During the training, we ran 250 epochs of all the training examples. For each epoch, we randomized the training examples and conducted a mini-batch training with a batch size of 128 ($m=128$).

In this paper, we experimented with three window sizes: 3, 5 and 7, each of which has 400 filters. Every filter performs convolution on the sentence matrix and generates variable-length feature maps. We got the best results using the single window of size 3 (see Section~\ref{sec:results and discussion})

\section{Results and Discussion}
\label{sec:results}

\subsection{Data}

We evaluated \mdcnn on two benchmarking PPI corpora, AIMed~\cite{bunescu2005comparative} and BioInfer~\cite{pyysalo2007bioinfer}. These two corpora have different sizes (Table~\ref{tab:corpus}) and vary slightly in their definition of PPI~\cite{pyysalo2008comparative}.
\begin{table}
\caption{Statistics of the corpora.\label{tab:corpus}}
\centering
\begin{tabularx}{.48\textwidth}{Xrrr}
\hline
Corpus & Sentences & \# Positives & \# Negatives\\
\hline
AIMed & 1,955 & 1,000 & 4,834\\
BioInfer & 1,100 & 2,534 & 7,132\\
\hline
\end{tabularx}
\end{table}

\citet{tikk2010comprehensive} conducted a comparison of a variety of PPI extraction systems on these two corpora\footnote{\url{http://mars.cs.utu.fi/PPICorpora}}. In order to compare, we followed their experimental setup to evaluate our methods: self-interactions were excluded from the corpora and 10-fold cross-validation (CV) was performed. 

\subsection{Results and discussion}
\label{sec:results and discussion}

Our system performance, as measured by Precision, Recall, and F1-score, is shown in Table~\ref{tab:evaluation}. To compare, we also include the results published in~\cite{tikk2010comprehensive, peng2015extended, vanlandeghem2008extracting, fundel2007relex}. Row 2 reports the results of the previous best deep learning system on these two corpora. Rows 3 and 4 report the results of two previous best single kernel-based methods, an APG kernel~\cite{airola2008all, tikk2010comprehensive} and an edit kernel~\cite{peng2015extended}. Rows 5-6 report the results of two rule-based systems. 
As can be seen, \mdcnn achieved the highest results in both precision and overall F1-score on both datasets. 

\begin{table*}
\newcommand{\rowno}[1]{$_\mathit{#1}$}
\caption{Evaluation results. Performance is reported in terms of Precision, Recall, and F1-score.\label{tab:evaluation}}
\centering
\begin{tabularx}{\textwidth}{lXrrrrrrr}
\hline
& & \multicolumn{3}{c}{AIMed} && \multicolumn{3}{c}{BioInfer}\\
\cline{3-5}\cline{7-9}
\multicolumn{2}{l}{Method} & P & R & F && P & R & F\\
\hline
\rowno{1} & \mdcnn & 67.3 & 60.1 & \textbf{63.5} && 62.7 & 68.2 & \textbf{65.3}\\
\rowno{2} & Deep neutral network~
\cite{zhao2016protein} & 51.5 & 63.4 & 56.1 && 53.9 & 72.9 & 61.6\\
\rowno{3} & All-path graph kernel~\cite{tikk2010comprehensive} & 49.2 & 64.6 & 55.3 && 53.3 & 70.1 & 60.0\\
\rowno{4} & Edit kernel~\cite{peng2015extended} & 65.3 & 57.3 & 61.1 && 59.9 & 57.6 & 58.7\\
\rowno{5} & Rich-feature~\cite{vanlandeghem2008extracting} & 49.0 & 44.0 & 46.0 && -- & -- & --\\
\rowno{6} & RelEx~\cite{fundel2007relex} & 40.0 & 50.0 & 44.0 && 39.0 & 45.0 & 41.0\\
\hline
%\rowno{7} & SVM-CW~\cite{miwa2009rich} & -- & -- & 64.0 && -- & -- & 66.7\\
%\hline
\end{tabularx}
\end{table*}

Note that we did not compare our results with two recent deep-learning approaches~\cite{hua2016shortest, quan2016multichannel}. This is because unlike other previous studies, they artificially removed sentences that cannot be parsed and discarded pairs which are in a coordinate structure. Thus, our results are not directly comparable with theirs.
Neither did we compare our method with~\cite{miwa2009rich} because they combined, in a rich vector, analysis from different parsers and the output of multiple kernels. 
%
%We did not compare our results with the assembly systems reported in~\cite{miwa2009rich} either, because they used a rich feature vector, combining analysis from different parsers and the values obtained from multiple kernels. 

To further test the generalizability of our method, we conducted the cross-corpus experiments where we trained the model on one corpus and tested it on the other (Table~\ref{tab:cc}). Here we compared our results with the shallow linguistic model which is reported as the best kernel-based method in~\cite{tikk2013detailed}. 

The cross-corpus results show that \mdcnn achieved 24.4\% improvement in F-score when trained on BioInfer and tested on AIMed, and 18.2\% vice versa. 
\begin{table*}
\caption{Cross-corpus results. Performance is reported in terms of Precision, Recall, and F1-score.\label{tab:cc}}
\centering
\begin{tabularx}{\textwidth}{lXrrrrrrr}
\hline
 & & \multicolumn{3}{c}{AIMed} && \multicolumn{3}{c}{BioInfer}\\
\cline{3-5}\cline{7-9}
Method & Training corpus & P & R & F && P & R & F\\
\hline
\mdcnn & AIMed & -- & -- & -- && 39.5 & 61.4 & \textbf{48.0}\\
 & BioInfer & 40.1 & 65.9 & \textbf{49.9} && -- & -- & --\\
Shallow linguistic~\cite{tikk2010comprehensive} & AIMed & -- & -- & -- && 29.2 & 66.8 & 40.6\\
& BioInfer & 76.8 & 27.2 & 41.5 && -- & -- & --\\
\hline
\end{tabularx}
\end{table*}

To better understand the advantages of \mdcnn over kernel-based methods, we followed the lead of~\cite{tikk2013detailed} to compare the method performance on some known ``difficult'' instances in AIMed and BioInfer. This subset of difficult instances is defined as ~10\% of all pairs with the least number of 14 kernels being able to classify correctly (Table~\ref{tab:difficult}). 
\begin{table}
\caption{Instances that are the most difficult to classify correctly by the collection of kernels using cross-validation~\cite{tikk2013detailed}.\label{tab:difficult}}
\centering
\begin{tabularx}{.48\textwidth}{Xrr}
\hline
Corpus & Positive difficult & Negative difficult\\
\hline
AIMed & 61 & 184\\
BioInfer & 111 & 295\\
\hline
\end{tabularx}
\end{table}

Table~\ref{tab:comparisons difficult} shows the comparisons between \mdcnn and kernel-based methods on difficult instances. The results of \mdcnn were obtained from the difficult instances combined from AIMed and BioInfer (172 positives and 479 negatives). And the results of APG, Edit, and SL were obtained from AIMed, BioInfer, HPRD50, IEPA, and LLL (190 positives and 521 negatives)~\cite{tikk2013detailed}. While the input datasets are different, our outcomes are remarkably higher than the prior studies. The results show that \mdcnn achieves 17.3\% in F1-score on difficult instances – which is more than three times better than other kernels. Since there are no examples of difficult instances that could not be classified correctly by at least one of the 14 kernel methods, below, we only list some examples that \mdcnn can classify correctly.
\begin{enumerate}
	\item Immunoprecipitation experiments further reveal that the fully assembled receptor complex is composed of two \textbf{IL-6}$_{\text{PROT1}}$, two \textbf{IL-6R alpha}$_{\text{PROT2}}$, and two gp130 molecules.
	\item The phagocyte NADPH oxidase is a complex of membrane cytochrome b558 (comprised of subunits p22-phox and gp91-phox) and three cytosol proteins (\textbf{p47-phox}$_{\text{PROT1}}$, p67-phox, and p21rac) that translocate to membrane and bind to \textbf{cytochrome b558}$_{\text{PROT2}}$.
\end{enumerate}
Together  with the conclusions in~\cite{tikk2013detailed}, ``positive pairs are more difficult to classify in longer sentences'' and ``most of the analyzed classifiers fail to capture the characteristics of rare positive pairs in longer sentences'', this comparison suggests that \mdcnn is probably capable of better capturing long distance features from the sentence and are more generalizable than kernel methods.
\begin{table}
\caption{Comparisons on the difficult instances with CV evaluation. Performance is reported in terms of Precision, Recall, and F1-score$^*$.\label{tab:comparisons difficult}}
%
%\vspace{.5ex}
\centering
\begin{tabularx}{.48\textwidth}{Xrrr}
\hline
Method & P & R & F\\
\hline
\mdcnn& 14.0& 22.7& \textbf{17.3}\\
All-path graph kernel& 4.3& 7.9& 5.5\\
Edit kernel& 4.8& 5.8& 5.3\\
Shallow linguistic & 3.6& 7.9&4.9\\
\hline
\end{tabularx}
\raggedright
\small $^*$~The results of McDepCNN were obtained on the difficult instances combined from AIMed and BioInfer (172 positives and 479 negatives). The results of others~\cite{tikk2013detailed} were obtained from AIMed, BioInfer, HPRD50, IEPA, and LLL (190 positives and 521 negatives).
\end{table}

Finally, Table~\ref{tab:comparisons} compares the effects of different parts in \mdcnn. Here we tested \mdcnn using 10-fold of AIMed. Row 1 used a single window with the length of 3, row 2 used two windows, and row 3 used three windows. The reduced performance indicate that adding more windows did not improve the model. This is partially because the multichannel in \mdcnn has captured good context features for PPI. Second, we used the single channel and retrained the model with window size 3. The performance then dropped 1.1\%. The results underscore the effectiveness of using the head word as a separate channel in CNN.
\begin{table}
\caption{Contributions of different parts in \mdcnn. Performance is reported in terms of Precision, Recall, and F1-score.\label{tab:comparisons}}
\centering
\begin{tabularx}{.48\textwidth}{Xrrrr}
\hline
Method & P & R & F & $\Delta$\\
\hline
window = 3 & 67.3 & 60.1 & 63.5 & \\
window = [3,5] & 60.9 & 62.4 & 61.6 & (1.9)\\
window = [3,5,7] & 61.7 & 61.9 & 61.8 & (1.7)\\
Single channel & 62.8 & 62.3 & 62.6 & (1.1)\\
\hline
\end{tabularx}
\end{table}

\section{Conclusion}

In this paper, we describe a multichannel dependency-based convolutional neural network for the sentence-based PPI task. Experiments on two benchmarking corpora demonstrate that the proposed model outperformed the current deep learning model and single feature-based or kernel-based models. Further analysis suggests that our model is substantially more generalizable across different datasets. Utilizing the dependency structure of sentences as a separated channel also enables the model to capture global information more effectively. 

In the future, we would like to investigate how to assemble different resources into our model, similar to what has been done to rich-feature-based methods~\cite{miwa2009rich} where the current best performance was reported (F-score of 64.0\% (AIMed) and 66.7\% (BioInfer)). We are also interested in extending the method to PPIs beyond the sentence boundary. Finally, we would like to test and generalize this approach to other biomedical relations such as chemical-disease relations~\cite{wei2016assessing}. 

\section*{Acknowledgments}

This work was supported by the Intramural Research Programs of the National Institutes of Health, National Library of Medicine. We are also grateful to Robert Leaman for the helpful discussion.

% include your own bib file like this:
%\bibliographystyle{acl}
\bibliography{acl2017}
%\bibliography{pbibliography}
\bibliographystyle{acl_natbib}

\end{document}